\documentclass[sigconf]{acmart}
\renewcommand\footnotetextcopyrightpermission[1]{} % removes footnote with conference information in first column

\AtBeginDocument{%
  }

\setcopyright{acmcopyright}
\copyrightyear{2022}
\acmYear{2022}
\acmConference[PIC '22]{PIC 2022: Person in Context (workshop at ACM Multimedia 2022)}{October 10--14, 2022}{Lisbon, Portugal}
\begin{document}

\title{STVGFormer: Spatio-Temporal Video Grounding with Static-Dynamic Cross-Modal Understanding}

\author{Zihang Lin\textsuperscript{1}, Chaolei Tan\textsuperscript{1}, Jian-Fang Hu\textsuperscript{1,3,}\footnotemark[1], Zhi Jin\textsuperscript{1}, Tiancai Ye\textsuperscript{2}, Wei-Shi Zheng\textsuperscript{1}}

\affiliation{
 \institution{\textsuperscript{1}Sun Yat-sen University, Guangzhou, China  
 \quad\textsuperscript{2}Tencent, Guangzhou, China
}
\institution{\textsuperscript{3}Guangdong Province Key Laboratory of Information
Security Technology, Guangzhou, China}
 \country{}
 }
\email{{linzh59, tanchlei}@mail2.sysu.edu.cn, {hujf5, jinzh26}@mail.sysu.edu.cn, tiancaiye@tencent.com, wszheng@ieee.org}

\renewcommand{\shortauthors}{Zihang Lin et al.}

\begin{abstract}
In this technical report, we introduce our solution to human-centric spatio-temporal video grounding task. We propose a concise and effective framework named STVGFormer, which models spatio-temporal visual-linguistic dependencies with a static branch and a dynamic branch. The static branch performs cross-modal understanding in a single frame and learns to localize the target object spatially according to intra-frame visual cues like object appearances. The dynamic branch performs cross-modal understanding across multiple frames. It learns to predict the starting and ending time of the target moment according to dynamic visual cues like motions. Both the static and dynamic branches are designed as cross-modal transformers. We further design a novel static-dynamic interaction block to enable the static and dynamic branches to transfer useful and complementary information from each other, which is shown to be effective to improve the prediction on hard cases. Our proposed method achieved 39.6\% vIoU and won the first place in the HC-STVG track of the 4th Person in Context Challenge.
\end{abstract}


\begin{CCSXML}
<ccs2012>
  <concept>
      <concept_id>10010147.10010178.10010224.10010225.10010231</concept_id>
      <concept_desc>Computing methodologies~Visual content-based indexing and retrieval</concept_desc>
      <concept_significance>500</concept_significance>
      </concept>
  <concept>
      <concept_id>10010147.10010178.10010224.10010225.10010227</concept_id>
      <concept_desc>Computing methodologies~Scene understanding</concept_desc>
      <concept_significance>500</concept_significance>
      </concept>
 </ccs2012>
\end{CCSXML}

\ccsdesc[500]{Computing methodologies~Visual content-based indexing and retrieval}
\ccsdesc[500]{Computing methodologies~Scene understanding}

\keywords{spatio-temporal video grounding, cross-modal learning}
\begin{teaserfigure}
  \centering
  \includegraphics[width=0.9\linewidth]{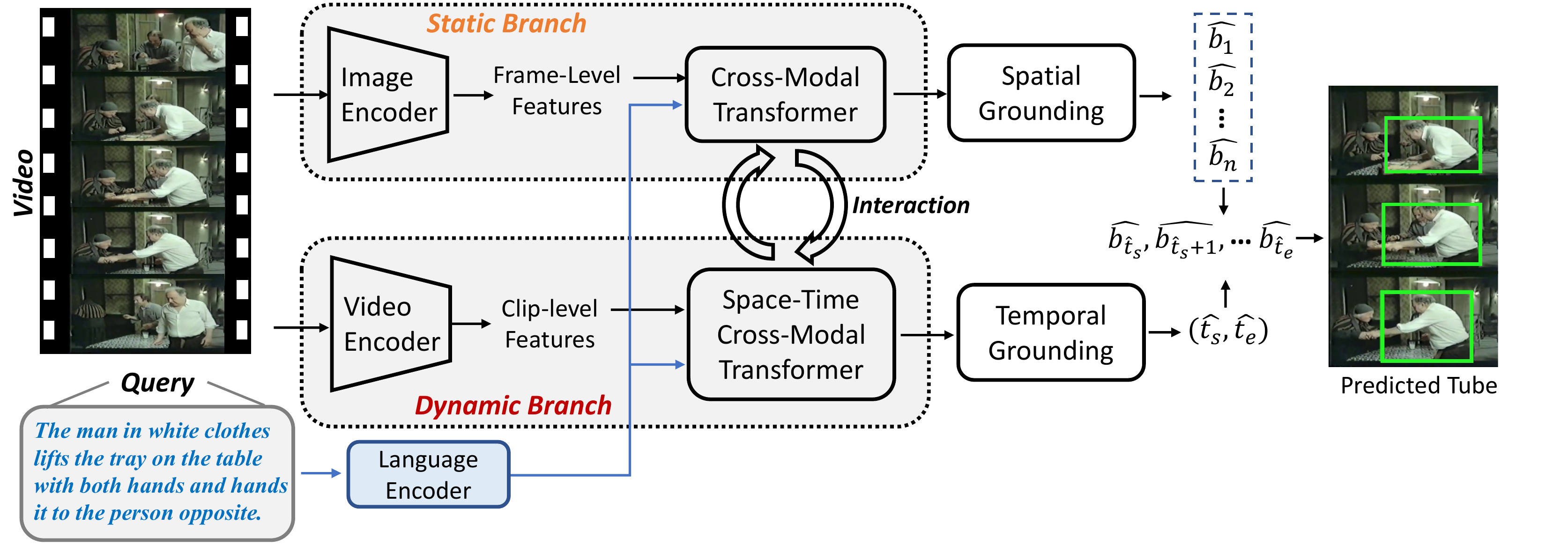}
  \caption{An overview of the proposed framework. Our framework mainly consists of a static branch and a dynamic branch. The static branch learns to predict the spatial location (i.e., bounding boxes $\hat{b_1}, \hat{b_2}, ..., \hat{b_n}$) of the target object according to static cues like human appearance. The dynamic branch learns to predict the starting and ending time $(\hat{t_s}, \hat{t_e})$ for the target moment according to dynamic cues like human action. We further devise a static-dynamic interaction block which enables the two branches to query useful and complementary information from the opposite branch.}
    \label{fig:framework}
\end{teaserfigure}

\maketitle

\renewcommand{\thefootnote}{\fnsymbol{footnote}}
\footnotetext[1]{Jian-Fang Hu is the corresponding author.}
\renewcommand{\thefootnote}{\arabic{footnote}}

\begin{figure*}[t]
  \centering
  \includegraphics[width=0.95\linewidth]{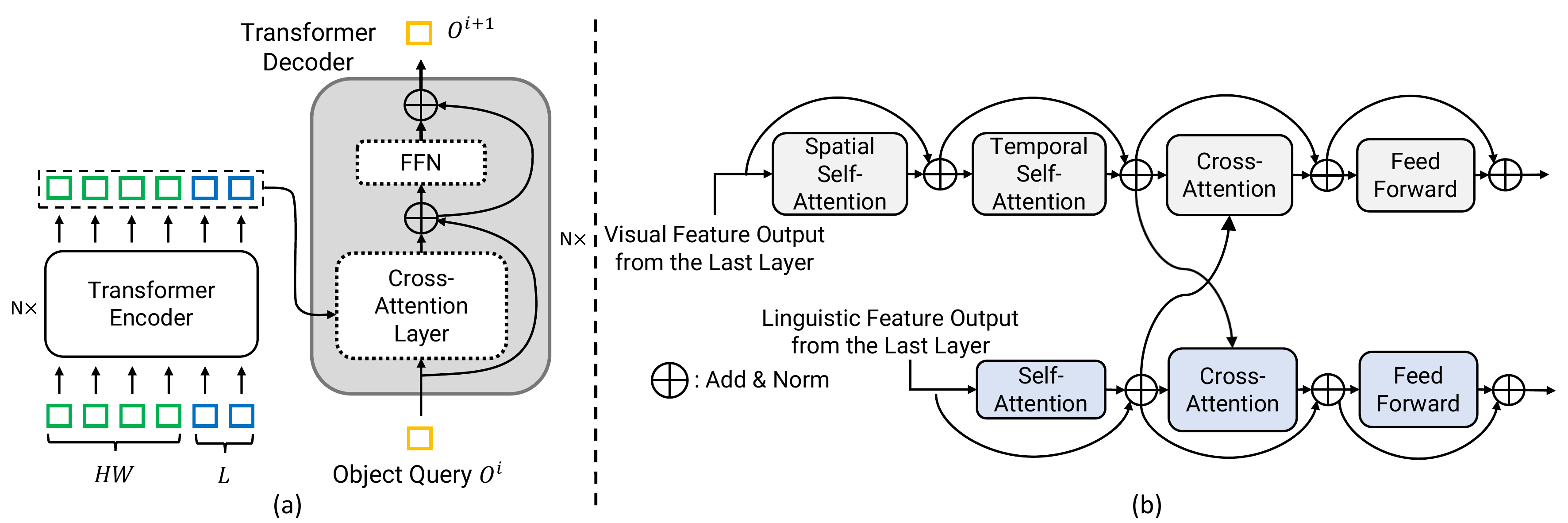}
  \caption{The detailed architectures of the proposed framework. (a): The cross-modal transformer in the static branch. (b): The architecture of each space-time cross-modal transformer layer in the dynamic branch.}
    \label{fig:detail_arch}
\end{figure*}

\section{Method}
Human-centric spatio-temporal video grounding (HCSTVG) task [5] aims to localize the target person spatially and temporally according to a language query. We propose a framework named STVGFormer to tackle this problem. As illustrated in Figure \ref{fig:framework}, our framework mainly consists of a static branch and a dynamic branch to model static and dynamic visual-linguistic dependencies for complete cross-modal understanding. The static branch performs cross-modal understanding for static contextual information, i.e., finding the target person that matches the query text in a still frame according to static visual cues, like appearance which is important for achieving accurate spatial grounding. The dynamic branch performs cross-modal understanding for dynamic contextual information, i.e., finding the temporal moment that best matches the query sentence according to dynamic visual cues, like action which is important for achieving accurate temporal grounding. In order to enable information transition between the two branches, we further develop a novel static-dynamic interaction block to exchange complementary information between the static and dynamic branches. In this way, both the static and dynamic branch can absorb useful and complementary information from the other one, which can help greatly reduce the uncertainty in the ambiguous and hard cases. Finally, we employ a bounding box prediction head on top of the static branch and a temporal moment prediction head on top of the dynamic branch to predict spatial and temporal grounding result, respectively. Overall, our framework is concise but effective. In the following, we will introduce each component in detail. 

\subsection{Static Branch}
The static branch is employed to perform cross-modal static context understanding. Following the implementations in MDETR\cite{mdetr}, we define our static branch as a stack of $N$ cross-modal transformer encoder layers and $N$ cross-modal transformer decoder layers, as presented in Figure \ref{fig:framework} and \ref{fig:detail_arch} (a). The inputs of our cross-modal transformer encoder are the concatenation of $\mathbb{R}^{HW\times d}$-sized visual features and $\mathbb{R}^{L\times d}$-sized language features, where $H,W$ are the resolution of the visual features extracted from the static image frame, $L$ is the number of the text tokens in the input text query. The visual and language features are obtained by a pre-trained image encoder and language encoder\cite{roberta}, respectively. Here, we also employ an FC embedding layer behind each feature extractor, so that both extracted visual features and language features are projected to have the same channel dimension. The outputs of the last cross-modal transformer encoder layer form a $\mathbb{R}^{(HW+L)\times d}$-sized cross-modal memory $\mathcal{M}_{t}$, which captures rich interactions between intra-frame static visual cues and linguistic descriptions. Then the cross-modal memory is fed into a stack of $N$ transformer decoder layers, together with a learnable object query vector $O \in \mathbb{R}^{d}$, so that the static information depicted in the images and texts can be encoded in the outputs (query vector) by repeatedly querying memory $\mathcal{M}_{t}$.

\subsection{Dynamic Branch}
The dynamic branch performs cross-modal understanding for dynamic contextual information. As illustrated in Figure \ref{fig:framework}, we first extract clip-level visual features $F_{c} \in \mathbb{R}^{T\times H \times W \times c}$ from $T$ uniformly sampled video clips by a pretrained 3D-CNN\cite{slowfast} and employ an FC embedding layer to project the channel dimension from $c$ to $d$. Then the features after projection are fed into a Space-Time Cross-Modal Transformer (STCMT) which consists of $N$ layers to model the visual-linguistic dependencies from a dynamic perspective. The detailed architecture of each layer in STCMT is illustrated in Figure \ref{fig:detail_arch} (b). In each layer, we first perform intra-modality self-attention for the dynamic visual features and linguistic features. For visual features, in order to reduce computation cost, we follow TimesFormer\cite{TimesFormer} to split the spatio-temporal attention into separate attentions. Denoting the visual features after self-attention as $F_v \in \mathbb{R}^{T\times H\times W\times d}$ and the language features after self-attention as $F_l \in \mathbb{R}^{L\times d}$. We then perform cross-attention between $F_v$ and $F_l$ as: 
\begin{align}
  Q_v^{(h,w)} = W_{qv}F_v^{(h,w)}, K_v= W_{kv}\overline{F_v}, V_v=W_{vv}\overline{F_v}, \nonumber \\ \nonumber
  Q_l = W_{ql}F_l, K_l = W_{kl}F_l, V_l=W_{vl}F_l, \\
  \widetilde{F_v^{(h,w)}} = F_v^{(h,w)} + \text{Attention}(Q_v^{(h,w)}, K_l, V_l),  \\
  \widetilde{F_l} = F_l +  \text{Attention}(Q_l, K_v, V_v), \nonumber\\
  \text{Attention(Q,K,V)} = \text{softmax}(\frac{QK_T}{\sqrt{d_k}})V, \nonumber
\end{align}
where $W_{qv}, W_{kv}, W_{vv}, W_{ql}, W_{kl}, W_{vl}$ are learnable weighted matrices for computing queries, keys and values for the attention mechanism. $\overline{F_v} \in \mathbb{R}^{T\times d}$ is obtained by conducting mean pooling on $F_v$ along the spatial dimensions. $d_k$ is the dimension of the queries and keys. $F_v^{(h,w)} \in \mathbb{R}^{T\times d}$ indicates the visual feature at spatial position $(h,w)$. $\widetilde{F_v}, \widetilde{F_l}$ are the output visual and linguistic features after cross-attention, respectively. The cross-attention operation is designed in the way that rich interactions between the dynamic visual and linguistic features can be explored, which enables the model to learn a powerful cross-modal representation about the depicted dynamic cues. This is essential for grounding temporal moments. After computing the cross-attention between the visual features and linguistic features, we finally employ Feed-Forward Network (FFN) to process both features. In our dynamic branch, rich context is learned from the two modalities, and the visual dynamic features and the linguistic features are fused well.

\begin{figure*}[t]
  \centering
  \includegraphics[width=0.9\linewidth]{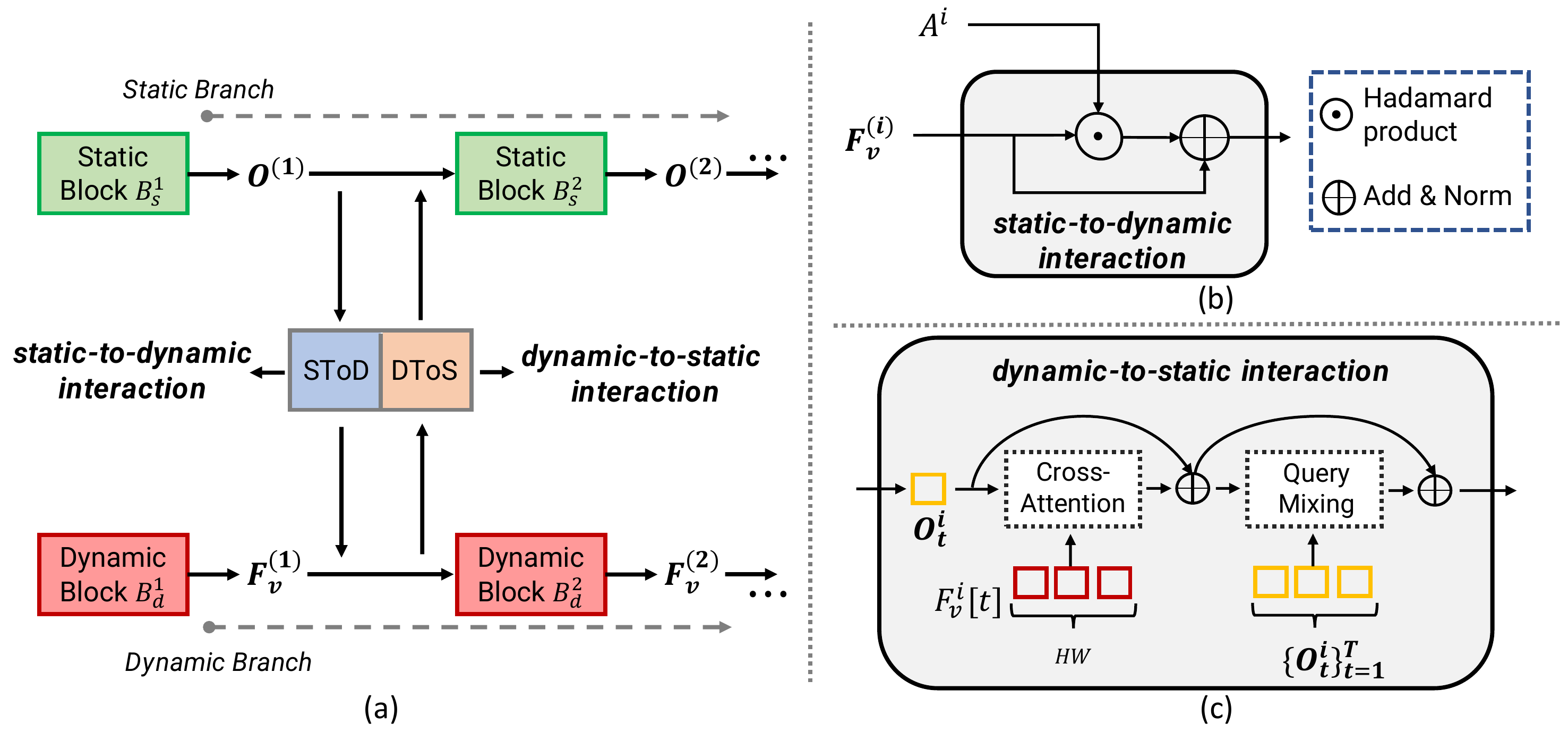}
  \caption{The architectures of the proposed Static-Dynamic Interaction Block.}
    \label{fig:cross_att}
\end{figure*}

\subsection{Static-Dynamic Interaction}
The Static-Dynamic Interaction Block (SDIB) enables information transition between the static branch and dynamic branch. As illustrated in Figure \ref{fig:cross_att}, this block is placed after each decoder layer $\left\{\mathcal{B}_s^{i}\right\}_{i=1}^{N}$ in static branch and each layer $\left\{\mathcal{B}_d^{i}\right\}_{i=1}^{N}$ in dynamic branch. It consists of a dynamic-to-static interaction block and a static-to-dynamic interaction block as illustrated in Figure \ref{fig:cross_att}. 

The static-to-dynamic interaction block is designed to guide the dynamic branch to attend to the image region that is highly related to the objects depicted in the query text, by utilizing the cross-attention matrices calculated in decoder layers of the static branch. Concretely, we employ static-to-dynamic interaction in each network layer as follows:
\begin{equation}
\widetilde{F^i_v} = \text{LayerNorm}(F^i_v + A^i \odot \text{FC}(F^i_v)),
\end{equation}
where $F^i_v \in \mathbb{R}^{T\times HW \times d}$ is the output visual feature of the \emph{i}-th layer in dynamic branch and $A^{i} \in \mathbb{R}^{T\times HW \times d}$ is the cross-attention weights (replicate to have $d$ channels at the last dimension) calculated between the object queries $\left\{O^i_t\right\}_{t=1}^{T}$ and corresponding encoded memories $\left\{M_t\right\}_{t=1}^{T}$ in $\mathcal{B}_s^{i}$. $\odot$ indicates hadamard product. The intuition behind this design is that the attention weights in the decoder layers of the static branch will attend to the object regions matching the language description, and this can serve as a strong guidance to help the dynamic branch focus more on the dynamic variations around the object-related regions. In this way, the model can better capture the motion of the target object. 

The dynamic-to-static interaction block is depicted in Figure \ref{fig:cross_att}(c). In each block, the object query $O^i_t$ in the static branch first queries some dynamic information from $F_v^i[t]$ (the output visual feature of the \emph{i}-th layer in Space-Time Cross-Modal Transformer for frame $t$). Then, the object queries of different frames are mixed up by a temporal self-attention layer. This block enhances the object query representation with cross-frame dynamic information like object motion and human action etc. With the proposed static-dynamic interaction block, both the static and dynamic branches can learn the complementary information from the other branch in an effective way.

\subsection{Prediction Heads}
In this section, we introduce the prediction heads used in our static branch and dynamic branch.

\noindent\textbf{Prediction Head for Static Branch}. The prediction head of the static branch is designed to predict the location of the target object. The input to the prediction head is the object query representation $O^{N}_t \in \mathbb{R}^{d}$ outputted by the static branch at $t$-th frame. We implement a 3-layer MLP to regress the bounding box location (represented by center coordinates and size) $\hat{b}_t \in \mathbb{R}^4$ of the target object. In order to make the static branch be aware of whether the input frame is well matched to the input text query, we further employ an FC layer to predict a score $\hat{p_t^{s}}$, which indicates whether frame $t$ is inside the target temporal moment.

\noindent\textbf{Prediction Head for Dynamic Branch}. The prediction head in the dynamic branch is designed to predict the time span of the target temporal moment. Specifically, we first perform mean pooling on the features outputted by the dynamic branch at the spatial dimension to obtain a temporal feature $F_d \in \mathbb{R}^{T\times d}$, then employ an FC layer to adjust the dimension from $d$ to $d_m$. We follow Aug2DTAN\cite{Aug2dTAN} to implement a 2D-proposal-based prediction head and construct the 2D proposal map $M\in \mathbb{R}^{T\times T\times d_m}$ as:
\begin{align}
  M_{ij} = \begin{cases} MeanPool\left(\left[F_d^{i}, F_d^{i+1},..., F_d^{j}\right]\right) & i \leq j \\
    \mathbf{0} & i > j
  \end{cases}
\end{align}
where $M_{ij}$ indicates the feature representation of temporal moment proposal $C_{ij}$ which consists of clips ${C_i, C_{i+1}, ..., C_{j}}$. Following the implementation in \cite{Aug2dTAN}, we employ several convolutional layers to the 2D map $M$ to obtain a score map $S \in \mathbb{R}^{T\times T\times 1}$ where each element $S_{ij}$ represents the matching score of the temporal proposal $C_{ij}$. For inference, we take the proposal with the highest score as the prediction of target temporal time span.

Similar to the static branch, we implement an auxiliary head (3-layer MLP) to predict whether each clip is inside the target temporal time span. Formally, $S_{aux} = MLP(F_d)$, where $S_{aux} \in \mathbb{R}^{T}$ and $S_{aux}^{i}$ indicate the probability of clip $i$ to be inside the target temporal moment.

\subsection{Model Training}
We train our model with loss $L=L_{s}+L_{d}$, where $L_s$ and $L_d$ are the losses on the output of the static branch and dynamic branches, respectively. Specifically, they are defined as following:
\begin{align}
 L_s &= \lambda_1 L_{l1}(\hat{b}, b)+\lambda_2 L_{gIoU}(\hat{b}, b) + \lambda_3 L_{aux}^{s},\\
 L_d &= \lambda_4 L_{tg}(\hat{S}, S)+ \lambda_5 L_{aux}^{d},
\end{align}
where $L_{l1}$ and $L_{gIoU}$ are L1 loss and gIoU loss on the predicted bounding boxes, respectively. $L_{tg}$ is a temporal grounding loss which is defined as a binary cross-entropy loss as done in \cite{2dtan, Aug2dTAN}, it is computed on the predicted score map $\hat{S}$ and the ground truth map $S$ in which each element is the $IoU$ between the corresponding proposal and ground truth temporal moment. $L_{aux}^{s}, L_{aux}^{d}$ are temporal attentive losses (implemented following \cite{tem_att_loss}) on the prediction of the auxiliary prediction head of the static branch and dynamic branch, respectively.

Specifically, we define the formulation of auxiliary losses as following:
\begin{align}
L_{aux}^{s} = \frac{-\sum_{t=1}^{T}m_t\log\hat{p_t^{s}}}{\sum_{t=1}^{T}m_t}, 
L_{aux}^{d} = \frac{-\sum_{t=1}^{T}m_t\log{S_{aux}^{t}}}{\sum_{t=1}^{T}m_t}.
\end{align}
where $m_t$ denotes the temporal mask at time $t$, i.e. $m_t$ takes 1 if $t\in [t_s, t_e]$ otherwise $m_t$ takes 0.

\section{Experiments}
\label{exp:all}
\subsection{Experimental Settings}
\paragraph{Evaluation Metrics}
We follow previous works\cite{Aug2dTAN, where-dose-it-exist, TubeDETR, STVGBert, HCSTVG_paper} to use mean vIoU as the main evaluation metric. vIoU is defined as $\frac{1}{\left|T_u\right|}\sum_{t\in T_i}IoU(\hat{b_t}, b_t)$, where $T_i$ and $T_u$ indicate the interaction and union between the time intervals obtained from ground truth annotation and system prediction, respectively. $\hat{b_t}, b_t$ are the predicted bounding box and ground truth bounding box for the $t$-th frame, respectively. We average the vIoU score over all the samples to obtain mean vIoU. We also report vIoU@R which indicates the proportion of samples with vIoU higher than $R$. 
\paragraph{Implementation Details}
We use ResNet101\cite{resnet} as our image encoder and Roberta-base\cite{roberta} as the text encoder to extract image visual features and language features. And we use the Slowfast\cite{slowfast} model pre-trained on AVA\cite{ava} as the video encoder. We initialize the weights of our static branch using pre-trained MDETR\cite{mdetr} as done in \cite{Aug2dTAN, TubeDETR}. We train our model with a batch size of 8 and the loss weights are set as $\lambda_1=5,\lambda_2=2, \lambda_3=0.5, \lambda_4=5, \lambda_5=1$. For details on training strategies, we follow the practice in MDETR\cite{mdetr}. To reduce GPU memory usage, we uniformly sampled 48 and 96 frames for the static branch during training and testing, respectively. During testing, we inference a video multiple times with different sampled frames until all frames are sampled at least once.

\subsection{Experimental Results}
%\paragraph{Main Results and Ablation Study.}
We first compare our method with state-of-the-arts on HCSTVG-v2 validation set\cite{HCSTVG_paper}. As shown in Table\ref{tab:HCSTVG2}, we outperform all previous methods by a considerable margin. Then we conduct ablation study by removing the static-to-dynamic interaction blocks (termed ``w/o s-to-d interaction''), the dynamic-to-static interaction blocks (termed ``w/o d-to-s interaction''), the whole static-dynamic interaction blocks (temred ``w/o interaction''). As shown in Table \ref{tab:HCSTVG2}, both the interaction blocks bring some performance gains, which demonstrates the effectiveness of the proposed static-dynamic interaction blocks for Spatio-Temporal Video Grounding.
\paragraph{HC-STVG Challenge.}
We submitted the prediction results of the proposed STVGFormer to the HC-STVG track of the 4th Person in Context Workshop. The naive STVGFormer achieved 36.8\% vIoU on the test set and the 5-model-ensemble STVGFormer achieved 39.6\% vIoU and won the first place in the HC-STVG track (results can be found on http://picdataset.com:8000/challenge/leaderboard/hcvg2022).

\begin{table}
  \centering
    \renewcommand{\arraystretch}{1.3}
    \caption{Comparison results on HCSTVG-v2\cite{HCSTVG_paper} validation set.}
    \setlength{\tabcolsep}{0.5mm}{
    \begin{tabular}{l|c|c|c}
        \toprule
        Method & m\_vIoU & vIoU@0.3 & vIoU@0.5\\
        \midrule
        Yu \textit{et al.}\cite{chengli2rd} & 30.0 & - & - \\
        MMN\cite{MMN} & 30.3 & 49.0 & 25.6\\
        Aug. 2D-TAN\cite{Aug2dTAN} & 30.4 & 50.4 & 18.8\\
        TubeDETR\cite{TubeDETR} & 36.4 & 58.8 & 30.6 \\
        \midrule
        STVGFormer(Ours) & \textbf{38.7} & \textbf{65.5} & \textbf{33.8} \\
        w/o s-to-d interaction & 37.1 & 62.3 & 30.2\\
        w/o d-to-s interaction & 37.5 & 63.6 & 31.4\\
        w/o interaction &  36.4 & 61.2 & 29.5\\
        \bottomrule
      \end{tabular}}
      \label{tab:HCSTVG2}
\end{table}

\begin{acks}
We would like to thank Bing Shuai for the helpful discussions.
\end{acks}

\bibliographystyle{ACM-Reference-Format}
\bibliography{ref.bib}


\begin{thebibliography}{15}

%%% the Web address.

\ifx \showCODEN    \undefined \def \showCODEN     #1{\unskip}     \fi
\ifx \showDOI      \undefined \def \showDOI       #1{#1}\fi
\ifx \showISBNx    \undefined \def \showISBNx     #1{\unskip}     \fi
\ifx \showISBNxiii \undefined \def \showISBNxiii  #1{\unskip}     \fi
\ifx \showISSN     \undefined \def \showISSN      #1{\unskip}     \fi
\ifx \showLCCN     \undefined \def \showLCCN      #1{\unskip}     \fi
\ifx \shownote     \undefined \def \shownote      #1{#1}          \fi
\ifx \showarticletitle \undefined \def \showarticletitle #1{#1}   \fi
\ifx \showURL      \undefined \def \showURL       {\relax}        \fi
% invisible to TeX
\providecommand\bibfield[2]{#2}
\providecommand\bibinfo[2]{#2}
\providecommand\natexlab[1]{#1}
\providecommand\showeprint[2][]{arXiv:#2}

\bibitem[Bertasius et~al\mbox{.}(2021)]%
        {TimesFormer}
\bibfield{author}{\bibinfo{person}{Gedas Bertasius}, \bibinfo{person}{Heng
  Wang}, {and} \bibinfo{person}{Lorenzo Torresani}.}
  \bibinfo{year}{2021}\natexlab{}.
\newblock \showarticletitle{Is Space-Time Attention All You Need for Video
  Understanding?}. In \bibinfo{booktitle}{\emph{Proceedings of the
  International Conference on Machine Learning (ICML)}}.
\newblock


\bibitem[Feichtenhofer et~al\mbox{.}(2019)]%
        {slowfast}
\bibfield{author}{\bibinfo{person}{Christoph Feichtenhofer},
  \bibinfo{person}{Haoqi Fan}, \bibinfo{person}{Jitendra Malik}, {and}
  \bibinfo{person}{Kaiming He}.} \bibinfo{year}{2019}\natexlab{}.
\newblock \showarticletitle{Slowfast networks for video recognition}. In
  \bibinfo{booktitle}{\emph{Proceedings of the IEEE/CVF international
  conference on computer vision}}. \bibinfo{pages}{6202--6211}.
\newblock


\bibitem[Gu et~al\mbox{.}(2018)]%
        {ava}
\bibfield{author}{\bibinfo{person}{Chunhui Gu}, \bibinfo{person}{Chen Sun},
  \bibinfo{person}{David~A Ross}, \bibinfo{person}{Carl Vondrick},
  \bibinfo{person}{Caroline Pantofaru}, \bibinfo{person}{Yeqing Li},
  \bibinfo{person}{Sudheendra Vijayanarasimhan}, \bibinfo{person}{George
  Toderici}, \bibinfo{person}{Susanna Ricco}, \bibinfo{person}{Rahul
  Sukthankar}, {et~al\mbox{.}}} \bibinfo{year}{2018}\natexlab{}.
\newblock \showarticletitle{Ava: A video dataset of spatio-temporally localized
  atomic visual actions}. In \bibinfo{booktitle}{\emph{Proceedings of the IEEE
  Conference on Computer Vision and Pattern Recognition}}.
  \bibinfo{pages}{6047--6056}.
\newblock


\bibitem[He et~al\mbox{.}(2016)]%
        {resnet}
\bibfield{author}{\bibinfo{person}{Kaiming He}, \bibinfo{person}{Xiangyu
  Zhang}, \bibinfo{person}{Shaoqing Ren}, {and} \bibinfo{person}{Jian Sun}.}
  \bibinfo{year}{2016}\natexlab{}.
\newblock \showarticletitle{Deep residual learning for image recognition}. In
  \bibinfo{booktitle}{\emph{Proceedings of the IEEE conference on computer
  vision and pattern recognition}}. \bibinfo{pages}{770--778}.
\newblock


\bibitem[Kamath et~al\mbox{.}(2021)]%
        {mdetr}
\bibfield{author}{\bibinfo{person}{Aishwarya Kamath}, \bibinfo{person}{Mannat
  Singh}, \bibinfo{person}{Yann LeCun}, \bibinfo{person}{Gabriel Synnaeve},
  \bibinfo{person}{Ishan Misra}, {and} \bibinfo{person}{Nicolas Carion}.}
  \bibinfo{year}{2021}\natexlab{}.
\newblock \showarticletitle{MDETR-modulated detection for end-to-end
  multi-modal understanding}. In \bibinfo{booktitle}{\emph{Proceedings of the
  IEEE/CVF International Conference on Computer Vision}}.
  \bibinfo{pages}{1780--1790}.
\newblock


\bibitem[Liu et~al\mbox{.}(2019)]%
        {roberta}
\bibfield{author}{\bibinfo{person}{Yinhan Liu}, \bibinfo{person}{Myle Ott},
  \bibinfo{person}{Naman Goyal}, \bibinfo{person}{Jingfei Du},
  \bibinfo{person}{Mandar Joshi}, \bibinfo{person}{Danqi Chen},
  \bibinfo{person}{Omer Levy}, \bibinfo{person}{Mike Lewis},
  \bibinfo{person}{Luke Zettlemoyer}, {and} \bibinfo{person}{Veselin
  Stoyanov}.} \bibinfo{year}{2019}\natexlab{}.
\newblock \showarticletitle{Roberta: A robustly optimized bert pretraining
  approach}.
\newblock \bibinfo{journal}{\emph{arXiv preprint arXiv:1907.11692}}
  (\bibinfo{year}{2019}).
\newblock


\bibitem[Su et~al\mbox{.}(2021)]%
        {STVGBert}
\bibfield{author}{\bibinfo{person}{Rui Su}, \bibinfo{person}{Qian Yu}, {and}
  \bibinfo{person}{Dong Xu}.} \bibinfo{year}{2021}\natexlab{}.
\newblock \showarticletitle{STVGBert: A Visual-Linguistic Transformer Based
  Framework for Spatio-Temporal Video Grounding}. In
  \bibinfo{booktitle}{\emph{Proceedings of the IEEE/CVF International
  Conference on Computer Vision (ICCV)}}. \bibinfo{pages}{1533--1542}.
\newblock


\bibitem[Tan et~al\mbox{.}(2021)]%
        {Aug2dTAN}
\bibfield{author}{\bibinfo{person}{Chaolei Tan}, \bibinfo{person}{Zihang Lin},
  \bibinfo{person}{Jian-Fang Hu}, \bibinfo{person}{Xiang Li}, {and}
  \bibinfo{person}{Wei-Shi Zheng}.} \bibinfo{year}{2021}\natexlab{}.
\newblock \showarticletitle{Augmented 2d-tan: A two-stage approach for
  human-centric spatio-temporal video grounding}.
\newblock \bibinfo{journal}{\emph{arXiv preprint arXiv:2106.10634}}
  (\bibinfo{year}{2021}).
\newblock


\bibitem[Tang et~al\mbox{.}(2021)]%
        {HCSTVG_paper}
\bibfield{author}{\bibinfo{person}{Zongheng Tang}, \bibinfo{person}{Yue Liao},
  \bibinfo{person}{Si Liu}, \bibinfo{person}{Guanbin Li},
  \bibinfo{person}{Xiaojie Jin}, \bibinfo{person}{Hongxu Jiang},
  \bibinfo{person}{Qian Yu}, {and} \bibinfo{person}{Dong Xu}.}
  \bibinfo{year}{2021}\natexlab{}.
\newblock \showarticletitle{Human-centric Spatio-Temporal Video Grounding With
  Visual Transformers}.
\newblock \bibinfo{journal}{\emph{IEEE Transactions on Circuits and Systems for
  Video Technology}} (\bibinfo{year}{2021}), \bibinfo{pages}{1--1}.
\newblock
\urldef\tempurl%
\url{https://doi.org/10.1109/TCSVT.2021.3085907}
\showDOI{\tempurl}


\bibitem[Wang et~al\mbox{.}(2021)]%
        {MMN}
\bibfield{author}{\bibinfo{person}{Zhenzhi Wang}, \bibinfo{person}{Limin Wang},
  \bibinfo{person}{Tao Wu}, \bibinfo{person}{Tianhao Li}, {and}
  \bibinfo{person}{Gangshan Wu}.} \bibinfo{year}{2021}\natexlab{}.
\newblock \showarticletitle{Negative Sample Matters: {A} Renaissance of Metric
  Learning for Temporal Grounding}.
\newblock \bibinfo{journal}{\emph{CoRR}}  \bibinfo{volume}{abs/2109.04872}
  (\bibinfo{year}{2021}).
\newblock


\bibitem[Yang et~al\mbox{.}(2022)]%
        {TubeDETR}
\bibfield{author}{\bibinfo{person}{Antoine Yang}, \bibinfo{person}{Antoine
  Miech}, \bibinfo{person}{Josef Sivic}, \bibinfo{person}{Ivan Laptev}, {and}
  \bibinfo{person}{Cordelia Schmid}.} \bibinfo{year}{2022}\natexlab{}.
\newblock \showarticletitle{TubeDETR: Spatio-Temporal Video Grounding with
  Transformers}. In \bibinfo{booktitle}{\emph{Proceedings of the IEEE/CVF
  Conference on Computer Vision and Pattern Recognition (CVPR)}}.
\newblock


\bibitem[Yu et~al\mbox{.}(2021)]%
        {chengli2rd}
\bibfield{author}{\bibinfo{person}{Yi Yu}, \bibinfo{person}{Xinying Wang},
  \bibinfo{person}{Wei Hu}, \bibinfo{person}{Xun Luo}, {and}
  \bibinfo{person}{Cheng Li}.} \bibinfo{year}{2021}\natexlab{}.
\newblock \showarticletitle{2rd Place Solutions in the HC-STVG track of Person
  in Context Challenge 2021}.
\newblock \bibinfo{journal}{\emph{arXiv preprint arXiv:2106.07166}}
  (\bibinfo{year}{2021}).
\newblock


\bibitem[Yuan et~al\mbox{.}(2019)]%
        {tem_att_loss}
\bibfield{author}{\bibinfo{person}{Yitian Yuan}, \bibinfo{person}{Tao Mei},
  {and} \bibinfo{person}{Wenwu Zhu}.} \bibinfo{year}{2019}\natexlab{}.
\newblock \showarticletitle{To find where you talk: Temporal sentence
  localization in video with attention based location regression}. In
  \bibinfo{booktitle}{\emph{Proceedings of the AAAI Conference on Artificial
  Intelligence}}, Vol.~\bibinfo{volume}{33}. \bibinfo{pages}{9159--9166}.
\newblock


\bibitem[Zhang et~al\mbox{.}(2020a)]%
        {2dtan}
\bibfield{author}{\bibinfo{person}{Songyang Zhang}, \bibinfo{person}{Houwen
  Peng}, \bibinfo{person}{Jianlong Fu}, {and} \bibinfo{person}{Jiebo Luo}.}
  \bibinfo{year}{2020}\natexlab{a}.
\newblock \showarticletitle{Learning 2d temporal adjacent networks for moment
  localization with natural language}. In \bibinfo{booktitle}{\emph{Proceedings
  of the AAAI Conference on Artificial Intelligence}},
  Vol.~\bibinfo{volume}{34}. \bibinfo{pages}{12870--12877}.
\newblock


\bibitem[Zhang et~al\mbox{.}(2020b)]%
        {where-dose-it-exist}
\bibfield{author}{\bibinfo{person}{Zhu Zhang}, \bibinfo{person}{Zhou Zhao},
  \bibinfo{person}{Yang Zhao}, \bibinfo{person}{Qi Wang},
  \bibinfo{person}{Huasheng Liu}, {and} \bibinfo{person}{Lianli Gao}.}
  \bibinfo{year}{2020}\natexlab{b}.
\newblock \showarticletitle{Where Does It Exist: Spatio-Temporal Video
  Grounding for Multi-Form Sentences}. In \bibinfo{booktitle}{\emph{Proceedings
  of the IEEE/CVF Conference on Computer Vision and Pattern Recognition
  (CVPR)}}.
\newblock


\end{thebibliography}

\end{document}